%% file: main.tex
\documentclass[a4paper, 10pt, conference]{ieeeconf}      % Use this line for a4
\usepackage{FG2023}

\usepackage[pagebackref=true,breaklinks=true,letterpaper=true,colorlinks,bookmarks=false]{hyperref}
\usepackage{amsmath,graphicx}

\usepackage{amsmath}
\usepackage{amssymb}
\usepackage{bm}

\usepackage{booktabs}

\usepackage[linesnumbered,ruled,vlined]{algorithm2e}
\SetKwInput{KwInput}{Input}
\SetKwInput{KwOutput}{Output}
\usepackage{commath}

\usepackage{caption}
\usepackage{subcaption}
\usepackage{verbatim}
\usepackage{gensymb}

\usepackage{rotating}
\usepackage{marvosym}
\usepackage{multirow}

\FGfinalcopy % *** Uncomment this line for the final submission

\IEEEoverridecommandlockouts\pubid{\makebox[\columnwidth]{979-8-3503-4544-5/23/\$31.00~\copyright{}2023 IEEE \hfill}
\hspace{\columnsep}\makebox[\columnwidth]{ }}

\def\FGPaperID{0026} % *** Enter the FG2023 Paper ID here

\title{\LARGE \bf Video Inference for Human Mesh Recovery with Vision Transformer}

\author{\parbox{16cm}{\centering
    {\large Hanbyel Cho$^1$, Jaesung Ahn$^2$, Yooshin Cho$^1$, Junmo Kim$^{1,2}$}\\
    {\normalsize
    $^1$ School of Electrical Engineering, KAIST, South Korea\\
    $^2$ Kim Jaechul Graduate School of AI, KAIST, South Korea}\\
    {\tt\small \texttt{\{\href{mailto:tlrl4658@kaist.ac.kr}{tlrl4658},\href{mailto:jaesung02@kaist.ac.kr}{jaesung02},\href{mailto:choys95@kaist.ac.kr}{choys95},\href{mailto:junmo.kim@kaist.ac.kr}{junmo.kim}\}@kaist.ac.kr}}}
}

\begin{document}

\ifFGfinal
\thispagestyle{empty}
\pagestyle{empty}
\else
\author{Anonymous FG2023 submission\\ Paper ID \FGPaperID \\}
\pagestyle{plain}
\fi
\maketitle

\newcommand{\etal}{\textit{et al}.}
\newcommand{\ie}{\textit{i}.\textit{e}.}
\newcommand{\eg}{\textit{e}.\textit{g}.}

%%%%%%%%% BODY INPUTS
\input{body_tex/0_Abstract}
\input{body_tex/1_Introduction}
\input{body_tex/2_Methodology}
\input{body_tex/3_Experiments}
\input{body_tex/4_Conclusion}

{\small
\bibliographystyle{ieee}
\bibliography{main}
}

\end{document}

%% file: body_tex/0_Abstract.tex
%%%%%%%%% ABSTRACT %%%%%%%%%
\begin{abstract}
    Human Mesh Recovery (HMR) from an image is a challenging problem because of the inherent ambiguity of the task. Existing HMR methods utilized either temporal information or kinematic relationships to achieve higher accuracy, but there is no method using both. Hence, we propose \emph{``Video Inference for \textbf{H}uman \textbf{M}esh \textbf{R}ecovery with \textbf{Vi}sion \textbf{T}ransformer (HMR-ViT)''} that can take into account both temporal and kinematic information. In HMR-ViT, a \emph{Temporal-kinematic Feature Image} is constructed using feature vectors obtained from video frames by an image encoder. When generating the feature image, we use a \emph{Channel Rearranging Matrix (CRM)} so that similar kinematic features could be located spatially close together. The feature image is then further encoded using \emph{Vision Transformer}, and the SMPL pose and shape parameters are finally inferred using a regression network. Extensive evaluation on the 3DPW and Human3.6M datasets indicates that our method achieves a competitive performance in HMR.
\end{abstract}

%% file: body_tex/1_Introduction.tex
%%%%%%%%% INTRODUCTION %%%%%%%%%
\section{Introduction}
    %[P1]
    Human Mesh Recovery (HMR)~\cite{ref15_icip_SMPLify,ref13_icip_HMR,ref14_icip_SPIN, ref18_icip_CMR, ref15_NeuralBodyFit, ref12_Coherent, ref_appearance, ref_texturepose, ref_coarse2fine, ref_delving, rw_ref8_BodyNet, rw_ref9_Self} is a problem that uses RGB inputs to infer the human body model (\eg, Skinned Multi-Person Linear model (SMPL)~\cite{ref1_icip_SMPL}, SMPL-X~\cite{ref22_icip_smplify-x}, STAR~\cite{ref25_icip_star}, and GHUM~\cite{ref2_icip_GHUM}) parameters that represents a person's three-dimensional (3D) pose and shapes. Along with the 3D joint-based approaches~\cite{ref7_martinez2017simple, ref8_zhao2019semantic, ref9_pavllo20193d, camdisthumanpose3d, ref10_cai2019exploiting, ref11_anatomy3D, ref17_DeepKinematics_CVPR2020, ref12_liu2020attention}, HMR is a fundamental task of computer vision, and is highly sought in downstream applications such as computer graphics, robotics, and AR/VR; however, it is challenging to achieve high accuracy owing to the inherent ambiguity (\eg, depth and occlusion) of the task.

    %[P2]
    Recently, to eliminate this ambiguity, many researchers utilized either temporal~\cite{ref3_icip_VIBE,ref17_icip_HMMR,ref4_icip_MEVA, ref16_icip_TCMR} or kinematic information~\cite{ref5_icip_METRO} in HMR. Kocabas~\etal~\cite{ref3_icip_VIBE} attempted to understand human behavior by encoding temporal information from video inputs, thereby overcoming depth ambiguity. Lin~\etal~\cite{ref5_icip_METRO} allowed the model to understand the non-local relationships between body joints through Masked Vertex Modeling by using Transformer~\cite{ref6_icip_Transformer}, and as a result, showed good performance in the case of occlusion exists. Although these methods have improved performance compared to the existing ones, there is still no method that takes advantage of both approaches.

    %[P3]
    To overcome this issue, in this study, we propose ``Video Inference for Human Mesh Recovery with Vision Transformer (\textbf{HMR-ViT})'' that can consider both temporal and kinematic information simultaneously. HMR-ViT mostly follows the framework of VIBE~\cite{ref3_icip_VIBE}, which first encodes each frame of a video sequence using the frozen pre-trained image encoder~\cite{ref13_icip_HMR, ref14_icip_SPIN} and then extracts the temporal information between each feature vector of the frame. However, unlike VIBE, which only performs temporal modeling using Gated Recurrent Units (GRUs)~\cite{ref7_icip_GRU}, our method exploits the network architecture of \emph{Vision Transformer (ViT)}~\cite{ref8_icip_ViT} to encode temporal and kinematic information simultaneously.

    %[P4]
    To achieve this, in HMR-ViT, a \emph{Temporal-kinematic Feature Image} is first constructed by concatenating the feature vectors generated by the image encoder from each frame along the time axis. The height of the constructed feature image denotes the time dimension and the width denotes the channel dimension of a feature vector extracted from each frame that can be considered to contain the kinematic information of a person. Then, we encode the information by considering the feature image as an image input of Vision Transformer. As in Dosovitskiy~\etal~\cite{ref8_icip_ViT}, the feature image is reshaped into a sequence of flattened 2D patches. Because each patch is composed of temporally and kinematically close information, our HMR-ViT can consider both temporal and kinematic information by modeling the relationship between these patches using an attention mechanism~\cite{ref6_icip_Transformer}.

    %[P5]
    In addition, we propose a learnable \emph{Channel Rearranging Matrix (CRM)} to further improve the performance of HMR-ViT by allowing spatially close kinematic features to be located close to each other on the channel dimension when generating the feature image. This allows each patch to be composed of information with a more similar kinematic meaning by sorting the width elements of the feature image. Finally, we use the regression network to infer the SMPL pose and shape parameters from the feature encoded by Vision Transformer. We conduct extensive evaluation on the 3DPW~\cite{ref9_icip_3dpw} and Human3.6M~\cite{ref23_icip_h36m} datasets, and the results indicate that the proposed method is effective for HMR task.

    \begin{figure*}[t]
        \includegraphics[width=\linewidth]{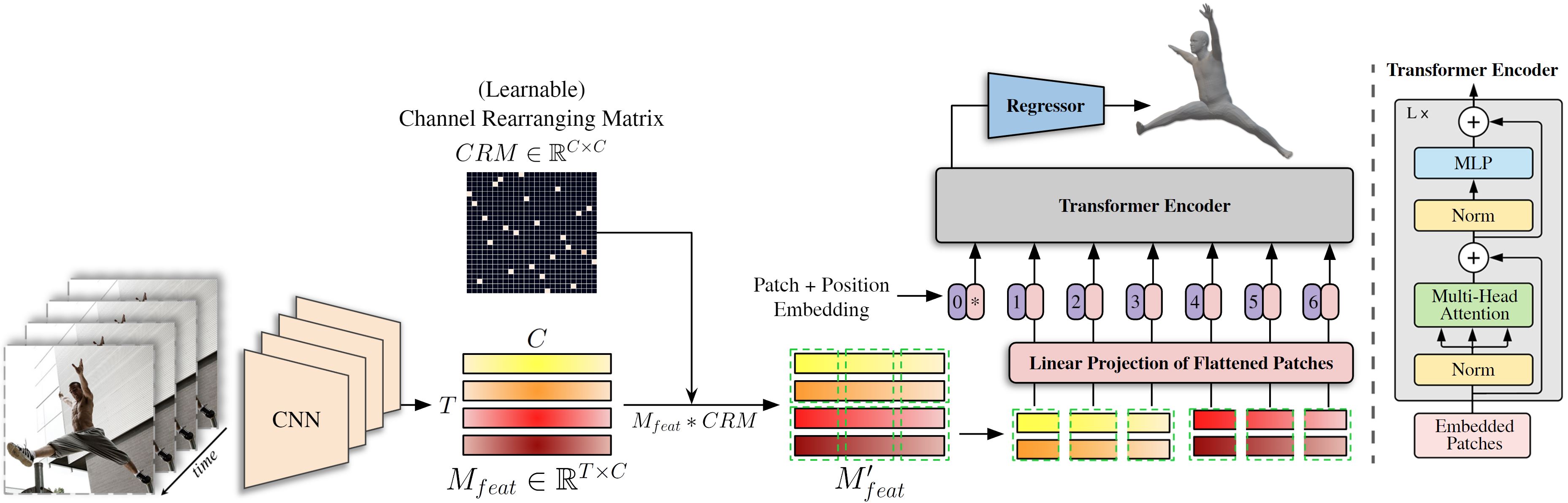}
        \caption{\textbf{Overview of the proposed Human Mesh Recovery with Vision Transformer (HMR-ViT).}
        Given a video of a person, HMR-ViT can understand the person's movement in the video by simultaneously modeling temporal and kinematic information using \emph{Vision Transformer} with the proposed \emph{Temporal-kinematic Feature Image} and \emph{Channel Rearranging Matrix}. Through this, our method achieves the more robust and consistent performance of human mesh recovery.
        }
        \label{fig:overall}
    \end{figure*}

    \vspace{2mm}
    In summary, our overall contribution is three-fold:
    %\vspace{-1mm}
    \begin{itemize}
        \item We propose a novel video-based HMR model named ``HMR-ViT'' that can take into account both temporal and kinematic information of a person in a video.
        \vspace{1mm}

        \item To achieve our goal, we propose the method to construct Temporal-kinematic Feature Image and make Channel Rearranging Matrix (CRM).
        \vspace{1mm}

        \item We confirm that our HMR-ViT successfully models both temporal and kinematic information and consequently outperforms the existing video-based HMR methods with improvement in computational efficiency.
    \end{itemize}

%% file: body_tex/2_Methodology.tex
%%%%%%%%% METHOD %%%%%%%%%
\vspace{1mm}
\section{Method}
The overall framework of our method is depicted in Fig.~\ref{fig:overall}. In this section, we provide a detailed description of HMR-ViT. We first provide a brief introduction to the SMPL human body model~\cite{ref1_icip_SMPL}. Subsequently, we explain the network architecture and training objectives of the proposed method.

\subsection{SMPL Body Model}
    Our method is built on top of SMPL~\cite{ref1_icip_SMPL}, which is a parametric human body model. The model provides a function $\mathcal{M}(\boldsymbol{\theta}, \boldsymbol{\beta})$ that outputs a human body mesh $B \in \mathbb{R}^{6890 \times 3}$ by using pose $\boldsymbol{\theta} \in \mathbb{R}^{72}$ and shape parameters $\boldsymbol{\beta} \in \mathbb{R}^{10}$ as inputs. The pose parameter consists of a relative 3D rotation of $23$ body joints and a global orientation in axis-angle representation. The shape parameter is the first $10$ coefficients of the PCA shape space, trained from thousands of registered human body scans. For a given mesh $B$, 3D joints $J$ are obtained using a pre-trained linear regressor $W$ as $J=WB$.

\subsection{Model Architecture}
    The goal of HMR-ViT is to infer human pose and shape from a video more robustly by simultaneously modeling \emph{temporal} and \emph{kinematic} information. For this, we incorporate Vision Transformer (ViT)~\cite{ref8_icip_ViT} into the video-based HMR~\cite{ref3_icip_VIBE}. In this section, HMR-ViT is described in detail in the following paragraphs: \emph{Temporal-kinematic Feature Image}, \emph{Channel Rearranging Matrix}, and \emph{Encoding with Vision Transformer}.

        \vspace{2mm}
        \noindent{\textbf{Temporal-kinematic Feature Image.}
        For a given input video \small$V = \{I_t\}_{t=1}^T$\normalsize\ of a single person with frame length $T$, we first encode each frame similar to the conventional video-based HMR method~\cite{ref3_icip_VIBE}. A feature vector $\mathbf{f}$ of each frame is obtained using the pre-trained CNN encoder $g$ (\ie, ResNet-50~\cite{ref12_icip_resnet}) as \small$\mathbf{f} = g(I) \in \mathbb{R}^{1 \times 1 \times C}$\normalsize, where the channel size \small$C$\normalsize\ is $2048$. Consequently, we obtained the set of the feature vectors \small$F = \{\mathbf{f}_t\}_{t=1}^T$\normalsize\ for the input video $V$.

        In order to model the temporal and kinematic information of $F$ simultaneously, we adopt ViT. Therefore, we construct a \emph{Temporal-kinematic Feature Image} (denoted as \small$M_{feat} \in \mathbb{R}^{T \times C}$\normalsize) by concatenating the feature vectors of $F$ along the time axis. The height and width of the constructed feature image $M_{feat}$ represent the temporal and kinematic information (represented as a channel component of the feature vector $\mathbf{f}$), respectively. From the feature image $M_{feat}$, our method can encode both types of information simultaneously by considering the feature image as an input 2D image of ViT.

        \vspace{2mm}
        \noindent{\textbf{Channel Rearranging Matrix.}
        To use a feature image $M_{feat}$ as a 2D image input of ViT, $M_{feat}$ should first be divided into multiple patches. When dividing the feature image into patches, information with similar temporal and kinematic meaning should be grouped to ensure good modeling between patches with different information.

        However, the width elements of the feature image $M_{feat}$ representing the channel components of the feature vectors $\mathbf{f}$, are not arranged in a kinematically meaningful order. Therefore, we propose \emph{Channel Rearranging Matrix} (denoted as \small$CRM \in \mathbb{R}^{C \times C}$\normalsize) that sorts the width elements of $M_{feat}$ such that spatially close kinematic features are located close to each other on the channel dimension. The $CRM$ matrix has a value of \emph{one} for only one element of each row and column and \emph{zeroes} for the rest. We implement the matrix by sequentially applying \emph{softmax} with \emph{temperature scaling} to the rows and columns of the randomly initialized trainable matrix. Finally, we multiply the feature image $M_{feat}$ by the $CRM$ matrix to obtain the refined feature image (denoted as \small$M'_{feat} \in \mathbb{R}^{T \times C}$\normalsize) in which the kinematic features are rearranged, as \small$M'_{feat} = M_{feat} * CRM$\normalsize, where $*$ denotes the matrix multiplication. Because the $CRM$ matrix is learnable, it is optimized during the training process.

        \vspace{2mm}
        \noindent\textbf{Encoding with Vision Transformer.}
        We have created a refined feature image $M'_{feat}$ whose height and width dimensions are well-arranged according to temporal and kinematic characteristics. To encode both temporal and kinematic information simultaneously, we use the refined feature map as an input 2D image of Vision Transformer. The refined feature image $M'_{feat}$ is first reshaped into a sequence of flattened 2D patches \small$\mathbf{m}_{patch} \in \mathbb{R}^{N \times (P_{t} \cdot P_{c})}$\normalsize, where $P_{t}$ and $P_{c}$ are the resolutions of each patch, and \small$N=(T/P_{t}) \cdot (C/P_{c})$\normalsize\ is the resulting number of patches, which is also considered as the input sequence length of Transformer. As in Vision Transformer, the sequence of patches is encoded through \emph{Linear Projection}, \emph{Position Embedding}, and \emph{Transformer Encoder}. We define the Vision Transformer model with only the classification head removed as $h$. From the sequence of patches $\mathbf{m}_{patch}$, the encoded feature vector $\mathbf{z}_{enc}$ is generated using the model $h$ as $\mathbf{z}_{enc} = h(\mathbf{m}_{patch}) \in \mathbb{R}^{2048}$.
        
        Finally, HMR-ViT infers the SMPL body model parameter $\Theta=\{\bm{\theta}, \bm{\beta}, \bm{\pi}\}$ from $\mathbf{z}_{enc}$ using the regressor network $\mathcal{R}(\cdot)$, where each component of $\Theta$ denotes the predicted pose, shape, and camera parameters. Our method infers parameter $\Theta$ for the middle frame of the input video sequence. We use weak-perspective camera model for the camera parameters $\bm{\pi} \in [s,t]$, where $s$ and $t$ denote the scale and translation parameters, respectively. From the inferred parameter $\Theta$, body mesh $B=\mathcal{M}(\bm{\theta}, \bm{\beta}) \in \mathbb{R}^{6890 \times 3}$ and 3D joints $J \in \mathbb{R}^{N_j\times 3}$ can be regressed, where $N_j$ denotes the number of joints. Furthermore, 2D keypoints $K \in \mathbb{R}^{N_j\times 2}$ can be obtained as $K =\bm{\Pi}(J)$, where $\bm{\Pi}(\cdot)$ denotes the weak-perspective camera projection function.

    \begin{figure}[t]
        \includegraphics[width=\columnwidth]{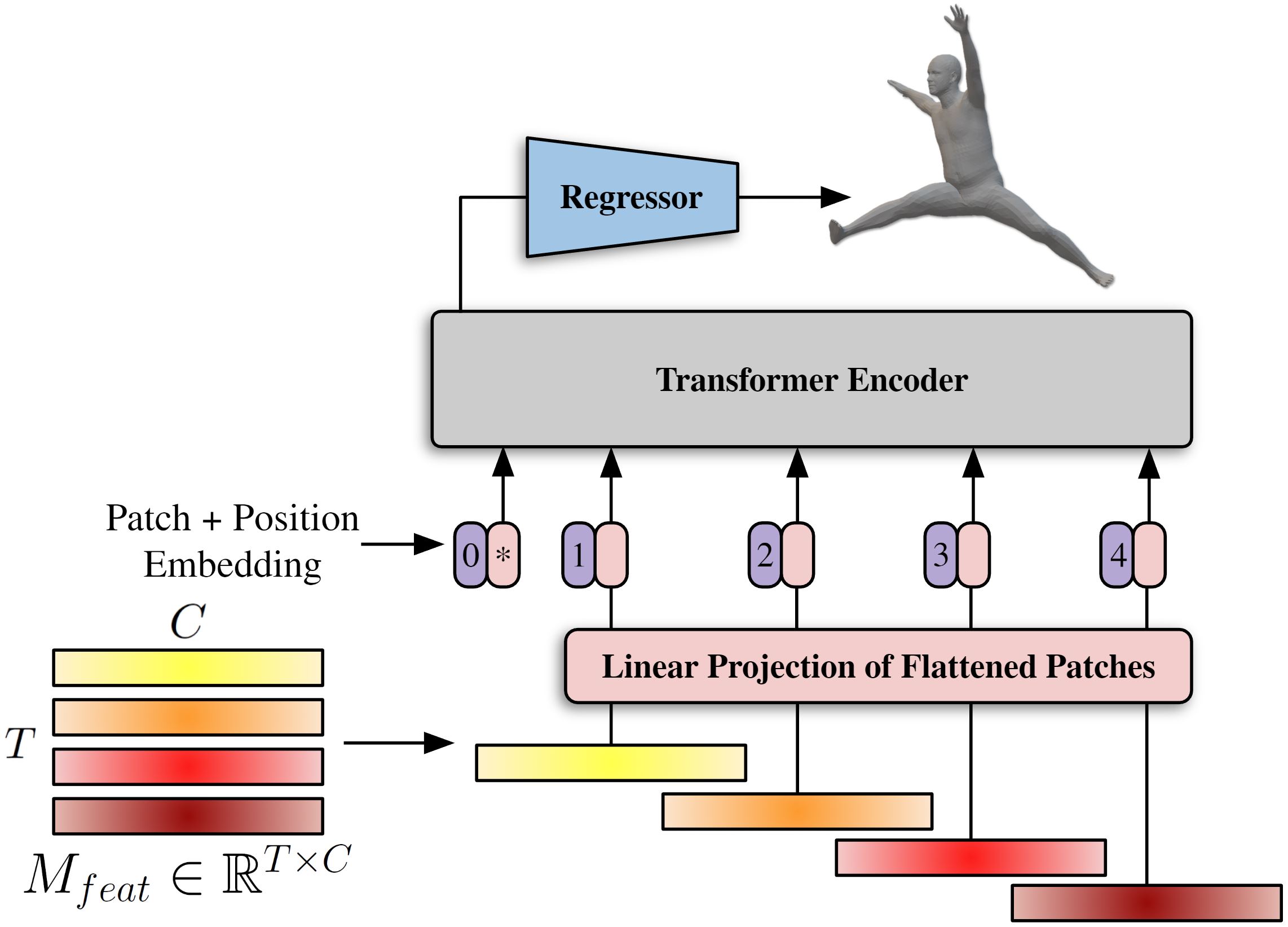}
        \caption{\textbf{Our baseline model that uses a naive Transformer.}
        We propose \emph{Our baseline} to which the Transformer is naively applied. It considers each feature vector extracted from an input video as an input token of the Transformer and encodes the feature \emph{without} modeling the kinematic information.
        }
        \label{fig:our_baseline}
    \end{figure}

    \subsection{Training Objective}
    To train the proposed model, we use the objective function commonly used in the conventional HMR paradigm~\cite{ref13_icip_HMR}. The loss is composed of 2D (\small$\mathcal{L}_{2D}$\normalsize), 3D (\small$\mathcal{L}_{3D}$\normalsize), SMPL pose (\small$\mathcal{L}_{pose}$\normalsize) and shape (\small$\mathcal{L}_{shape}$\normalsize) losses, each of which is \small$||\hat{K} - K||$\normalsize, \small$||\hat{J} - J||$\normalsize, \small$||\hat{\bm{\theta}} - \bm{\theta}||$\normalsize, and \small$||\hat{\bm{\beta}} - \bm{\beta}||$\normalsize, where $\hat{K}$, $\hat{J}$, $\hat{\bm{\theta}}$, and $\hat{\bm{\beta}}$ denote the predicted 2D keypoints, 3D joints, pose and shape parameters, respectively; $||\cdot||$ denotes the squared L2 norm. Moreover, we use the objective (\small$\mathcal{L}_{CRM}$\normalsize) such that the sum of each row and column of the \emph{CRM} matrix should be \emph{one} so that the matrix is trained with an appropriate sorting matrix. Overall, our total loss function is \small$\mathcal{L}_{total} = \lambda_{2D}\cdot\mathcal{L}_{2D} + \lambda_{3D}\cdot\mathcal{L}_{3D} + \lambda_{pose}\cdot\mathcal{L}_{pose} + \lambda_{shape}\cdot\mathcal{L}_{shape} + \lambda_{CRM}\cdot\mathcal{L}_{CRM}$\normalsize, where the \small$\lambda_{(\cdot)}$\normalsize\ denotes the hyperparameter for each loss function. We use each loss function when related data are available. Additionally, we report the results of adding the motion compensation constraint using AMASS dataset to our HMR-ViT, as mentioned in Table~\ref{tab:sota_quant}, for fair comparison with SOTA~\cite{ref3_icip_VIBE}. In this case, the motion compensation constraint is applied using the same discriminator as in \cite{ref3_icip_VIBE}.

%% file: body_tex/3_Experiments.tex
%%%%%%%%% EXPERIMENTS %%%%%%%%%
\vspace{1mm}
\section{Experiments}
    \subsection{Implementation Details}
        For the CNN encoder $g$, we adopt the ResNet-50~\cite{ref12_icip_resnet} model pre-trained with a single-image human mesh recovery task~\cite{ref13_icip_HMR,ref14_icip_SPIN}. As in Kocabas~\etal~\cite{ref3_icip_VIBE}, we precompute the feature vector from all datasets, and the CNN encoder $g$ is frozen during the training process. For Vision Transformer model, we use the same model architecture proposed in Dosovitskiy~\etal~\cite{ref8_icip_ViT}, except for the presence of a classification head and the number of layers in Transformer Encoder. We use the same regressor network $\mathcal{R}$ as the model used by Kolotouros~\etal~\cite{ref14_icip_SPIN}. We initialize $\mathcal{R}$ with the pretrained weights provided by \cite{ref14_icip_SPIN}. The maximum iteration number of the regressor is three, as in~\cite{ref13_icip_HMR,ref14_icip_SPIN,ref3_icip_VIBE}. For the training process, we adopt Adam~\cite{ref24_icip_adam} with batch size of $32$ as our optimizer. We set the hyperparameters $\lambda_{2D}$, $\lambda_{3D}$, $\lambda_{pose}$, $\lambda_{shape}$, and $\lambda_{CRM}$ to $300$, $300$, $60$, $0.06$, and $1$, respectively. We perform the training with $300$ epochs and use $5e{-}5$ for the learning rate.

    \begin{figure}[t]
        \includegraphics[width=\columnwidth]{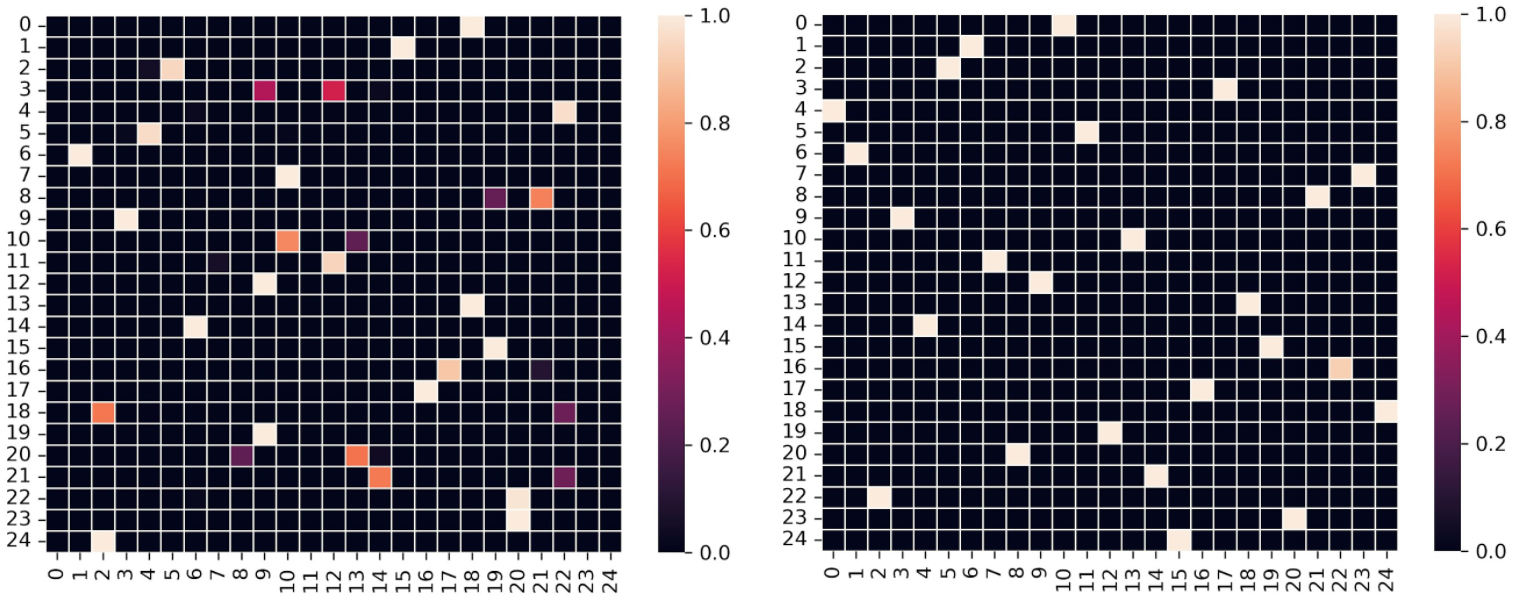}
        \caption{\textbf{Convergence of the Channel Rearranging Matrix.} We report the convergence result of the CRM matrix trained with the constraints $\mathcal{L}_{CRM}$ when the channel size $C$ is $25$. The CRM matrix converges from a randomly initialized matrix (\emph{\textbf{left}}) to an appropriate sorting matrix (\emph{\textbf{right}}).
        }
        \label{fig:crm}
    \end{figure}

    \subsection{Datasets and Evaluation Metrics}
        \vspace{1mm}
        \noindent\textbf{Datasets.}
        We use a mixture of 2D and 3D datasets in previous studies~\cite{ref13_icip_HMR,ref3_icip_VIBE,ref4_icip_MEVA,ref17_icip_HMMR}. For training, we use InstaVariety~\cite{ref17_icip_HMMR}, PoseTrack~\cite{ref19_icip_posetrack}, and PennAction~\cite{ref20_icip_penn} as the 2D datasets. PoseTrack and PennAction datasets have ground-truth 2D keypoint labels, and the InstaVariety has pseudo ground-truth 2D keypoint labels estimated by 2D keypoint estimator~\cite{ref_openpose}. MPI-INF-3DHP~\cite{ref21_icip_mpii3d} and Human3.6M~\cite{ref23_icip_h36m} are used as the 3D datasets for training. The pseudo-ground truth SMPL label is obtained using Pavlakos~\etal~\cite{ref22_icip_smplify-x} and Kolotouros~\etal~\cite{ref14_icip_SPIN}. For the evaluation, the 3DPW and Human3.6M datasets are used.

        \vspace{2mm}
        \noindent\textbf{Evaluation metrics.}
        We report the performance of HMR-ViT on PVE (per-vertex error), MPJPE (mean per joint position error), and PA-MPJPE (mean per joint position error after Procrustes-alignment. More specifically, PVE (per-vertex error) is the error of summing the Euclidean distance between the inferred mesh vertexes and corresponding vertexes of ground truth mesh, and the lower the value, the better the reconstruction quality of the body surface. MPJPE (mean per joint position error) is an error for 3D joints regressed from SMPL body mesh and this joint-based metric can estimate the restoration accuracy for only poses except body shapes. Finally, PA-MPJPE (MPJPE after Procrustes-aligned) performs Procrustes-alignment(statistical shape analysis) between the inferred 3D joints and the ground truth 3D joints, and then measures MPJPE, thus allowing us to know the restoration accuracy of the poses except for global orientation and scale difference.

    \subsection{Experimental Results}
        In this section, we verify the efficacy of the proposed method. First, we compare the performance of HMR-ViT with existing HMR methods both quantitatively and qualitatively. Then, we verify the effectiveness of each of the proposed methods. Finally, we perform an ablation study on the patch size.

    \begin{figure}[!t]
        \includegraphics[width=\columnwidth]{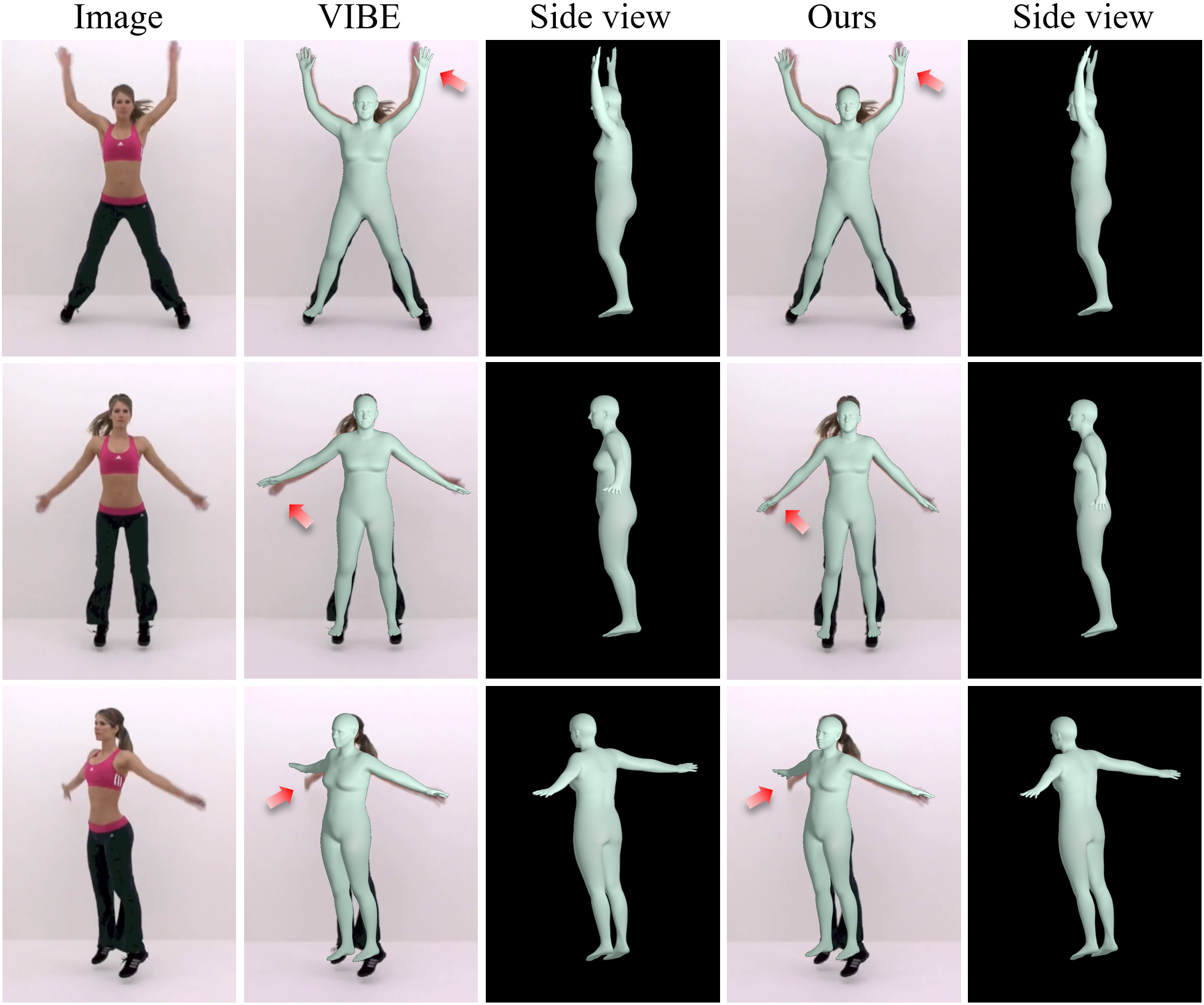}
        \caption{\textbf{Qualitative results.}
        Qualitative comparison of HMR-ViT (Ours) with VIBE~\cite{ref3_icip_VIBE} on the video of jumping person. The proposed method shows more plausible results.
        }
        \label{fig:qualitative}
    \end{figure}

    \begin{table}[!t]
        \centering
        \normalsize
        \resizebox{\columnwidth}{!}{
        \begin{tabular}{l|ccc}
        \toprule%($\downarrow$)
        \multicolumn{1}{c|}{\multirow{2}[4]{*}{Method}} & \multicolumn{3}{c}{3DPW} \\
        \cmidrule{2-4}
        \multicolumn{1}{c|}{} & PVE & MPJPE & PA-MPJPE \\
        \midrule
            Kanazawa~\etal~\cite{ref13_icip_HMR}    ~& -     & 130.0   & 76.7 \\
            Kolotouros~\etal~\cite{ref14_icip_SPIN} ~& 116.4 & 96.9  & 59.2 \\
        \midrule
            Kanazawa~\etal~\cite{ref17_icip_HMMR}   ~& 139.3 & 116.5 & 72.6 \\
            Kocabas~\etal~\cite{ref3_icip_VIBE}     ~& 113.4 & 93.5  & \textbf{56.5} \\
        \cmidrule{1-4}
            HMR-ViT                                 ~& \underline{112.0} & \underline{93.2} & 59.3 \\
            HMR-ViT \textit{w. motion disc.}        ~& \textbf{110.8} & \textbf{91.5} & \underline{58.4} \\
        \cmidrule{1-4}
            HMR-ViT \textit{w. 3DPW}                ~& 103.8 & 85.4 & 55.9 \\
        \bottomrule
        \end{tabular}%
        }
        \caption{\textbf{Quantitative results on the 3DPW dataset.}
        Values are in the scale of mm. Best in bold, second-best underlined. HMR-ViT and HMR-ViT \emph{w. 3DPW} denote the method trained \emph{w/o.} and \emph{w.} the 3DPW train set, respectively. HMR-ViT \emph{w. motion disc.} denotes our method trained with motion compensation constraints using motion discriminator with AMASS~\cite{ref26_amass} dataset.
        }
        \label{tab:sota_quant}
    \end{table}

    \begin{table}[!t]
        \centering
    	\normalsize
    	\tabcolsep=1mm
    	\resizebox{\columnwidth}{!}{
    		\begin{tabular}{@{}l|ccc@{}}
    		\toprule
    		~~Method ~&~ PVE ~&~ MPJPE ~&~ PA-MPJPE ~~\\
    		\midrule
    		~~Our baseline              ~&~ 114.2 ~&~ 94.2 ~&~ \textbf{58.0} ~~\\
    		\midrule
    		~~HMR-ViT w/o. \emph{CRM}   ~&~ \underline{113.3} ~&~ \underline{93.6} ~&~ \underline{59.2} ~~\\
        	~~HMR-ViT                   ~&~ \textbf{112.0} ~&~ \textbf{93.2} ~&~ 59.3 ~~\\
    		\bottomrule
    		\end{tabular}
    	}
    	\caption{\textbf{Effectiveness of each proposed method.}
    	The results are evaluated on the 3DPW dataset. Values are in the scale of mm. Best in bold, second-best underlined.
    	}
        \label{tbl:effectiveness}
    \end{table}

        \vspace{2mm}
        \noindent\textbf{Comparison with State-of-the-Art.}
        Table~\ref{tab:sota_quant} shows the quantitative results of HMR-ViT and existing HMR methods on the 3DPW dataset. As shown in the table, the proposed method shows better performance in PVE and MPJPE metrics for both the frame-based~\cite{ref13_icip_HMR,ref14_icip_SPIN} and temporal-based methods~\cite{ref17_icip_HMMR,ref3_icip_VIBE}. In particular, it can be confirmed that HMR-ViT shows competitive performance with VIBE~\cite{ref3_icip_VIBE}, the state-of-the-art video-based HMR. In addition, we compared the qualitative results of the proposed method and Kocabas~\etal~\cite{ref3_icip_VIBE}. As shown in Fig.~\ref{fig:qualitative}, The proposed method shows more plausible results than the comparative method.

        Kocabas~\etal~\cite{ref3_icip_VIBE} reported in Table~\ref{tab:sota_quant} performed adversarial training to infer more realistic SMPL values using an additional large-scale motion capture dataset named AMASS~\cite{ref26_amass} that was not used in our method. For a fair comparison, we applied the same motion compensation loss using discriminator network with AMASS dataset (without using additional parameters in inference phase) as VIBE~\cite{ref3_icip_VIBE} to HMR-ViT (denoted as HMR-ViT \textit{w. motion disc.} in Table~\ref{tab:sota_quant}). Our method using AMASS showed a performance of $110.8$mm, $91.5$mm, and $58.4$mm on PVE, MPJPE, and PA-MPJPE metrics, respectively. This is a significant performance improvement of $4$\%, $6$\%, and $8$\% compared to VIBE~\cite{ref3_icip_VIBE}.

        \vspace{2mm}
        \noindent\textbf{Ablation Studies.}
        To verify the effectiveness of each of the proposed methods, we compare the proposed method with \emph{Our baseline} (as shown in Fig.~\ref{fig:our_baseline}) to which Transformer~\cite{ref6_icip_Transformer} is naively applied. Table~\ref{tbl:effectiveness} shows the results. As can be seen from the table, our method shows better performance than \emph{Our baseline}, and there is an additional performance improvement when the CRM matrix is applied. This verifies that constructing a temporal-kinematic feature image and using it as an image input of ViT is an effective method.
        
        Also, as shown in Fig.~\ref{fig:crm}, the CRM matrix converges to an appropriate sorting matrix. Moreover, we performed an ablation study on the patch size dividing the feature images. As shown in Table~\ref{tbl:patch}, the lower errors are shown when $P_c$ (denoted the size of the channel dimension of the divided feature image) and $P_t$ (denoted the size of the time dimension of the divided feature image) have small values. This proves that dividing the feature image into patches helps in modeling temporal and kinematic information.

    \begin{table}[!t]
        \centering
    	\normalsize
    	\tabcolsep=1mm
    		\begin{tabular}{@{}l|ccc@{}}
    		\toprule
    		~ ~~~&~ $P_c=128$ ~&~ $P_c=512$ ~~\\
    		\midrule
    		~ $P_t=3$ ~~~&~ 64.7 ~&~ 65.4 ~~\\
    		\midrule
    		~ $P_t=5$ ~~~&~ 65.2 ~&~ 66.1 ~~\\
    		\bottomrule
    		\end{tabular}
    	\caption{\textbf{Ablation study on the patch size.}
    	The results are the performance on MPJPE for the Human3.6M dataset. Values are in the scale of mm. $P_c$ and $P_t$ denote the sizes of the channel dimension and time dimension of the divided feature-kinetic feature image, respectively.
    	}
        \label{tbl:patch}
    \end{table}

    \begin{table}[!t]
        \centering
    	\normalsize
    	\tabcolsep=1mm
    	\resizebox{\columnwidth}{!}{
    		\begin{tabular}{@{}l|ccc@{}}
    		\toprule
    		~ ~~~&~ VIBE~\cite{ref3_icip_VIBE} ~&~ HMR-ViT (Ours) ~~\\
    		\midrule
    		~ $\small\# \normalsize$ of parameters ~~~&~ 69M ~&~ 44M ~~\\
    		\bottomrule
    		\end{tabular}
    	}
    	\caption{\textbf{Computational complexity.}
    	Values in the table represent the number of parameters used in each of VIBE~\cite{ref3_icip_VIBE} and HMR-ViT (Ours).
    	}
        \label{tab:complexity}
    \end{table}

        \vspace{2mm}
        \noindent\textbf{Analysis about Computational Complexity.}
        For computational complexity (number of trainable parameters), HMR-ViT requires only $44$M parameters, $36$\% less than those of SOTA (VIBE~\cite{ref3_icip_VIBE}), $69$M, as shown in Table~\ref{tab:complexity}. Also, as shown ``HMR-ViT'' item in Table~\ref{tab:sota_quant}, HMR-ViT shows better performance without using AMASS dataset used by VIBE~\cite{ref3_icip_VIBE}. As such, the performance improvement achieved in situations using much fewer parameters and data is not marginal thus demonstrating the efficiency and efficacy of the proposed method, where both spatial and temporal information is modeled simultaneously using the Temporal-Kinematic Feature Image and ViT encoder. Furthermore, as shown in ``HMR-ViT \textit{w. motion disc.}'' item of Table~\ref{tab:sota_quant}, our method achieves a much greater performance improvement when applying the same motion compensation as in VIBE~\cite{ref3_icip_VIBE}.

%% file: body_tex/4_Conclusion.tex
%%%%%%%%% CONCLUSION %%%%%%%%%
\vspace{1mm}
\section{Conclusion}
We presented a video-based HMR method named HMR-ViT that can model temporal and kinematic information simultaneously. To this end, we incorporated Vision Transformer into the conventional video-based HMR. For given video frames, we construct a \emph{Temporal-kinematic Feature Image} and apply the proposed \emph{Channel Rearranging Matrix} to use it as an input for \emph{Vision Transformer}. The experimental results indicate that HMR-ViT achieves superior performance with the highly efficient model in terms of computational complexity compared to the existing HMR methods, and the ablation studies verify the efficacy of each of the proposed methods.

\noindent{\textbf{Acknowledgements} This work was conducted by Center for Applied Research in Artificial Intelligence (CARAI) grant funded by DAPA and ADD (UD190031RD).

%% file: main.bbl
\begin{thebibliography}{10}\itemsep=-1pt

\bibitem{ref19_icip_posetrack}
M.~Andriluka, U.~Iqbal, E.~Insafutdinov, L.~Pishchulin, A.~Milan, J.~Gall, and
  B.~Schiele.
\newblock Posetrack: A benchmark for human pose estimation and tracking.
\newblock In {\em CVPR}, 2018.

\bibitem{ref15_icip_SMPLify}
F.~Bogo, A.~Kanazawa, C.~Lassner, P.~V. Gehler, J.~Romero, and M.~J. Black.
\newblock Keep it smpl: Automatic estimation of 3d human pose and shape from a
  single image.
\newblock In {\em ECCV}, 2016.

\bibitem{ref10_cai2019exploiting}
Y.~Cai, L.~Ge, J.~Liu, J.~Cai, T.-J. Cham, J.~Yuan, and N.~M. Thalmann.
\newblock Exploiting spatial-temporal relationships for 3d pose estimation via
  graph convolutional networks.
\newblock In {\em Proceedings of the IEEE/CVF International Conference on
  Computer Vision}, pages 2272--2281, 2019.

\bibitem{ref_openpose}
Z.~Cao, T.~Simon, S.-E. Wei, and Y.~Sheikh.
\newblock Realtime multi-person 2d pose estimation using part affinity fields.
\newblock In {\em Proceedings of the IEEE Conference on Computer Vision and
  Pattern Recognition (CVPR)}, July 2017.

\bibitem{ref11_anatomy3D}
T.~Chen, C.~Fang, X.~Shen, Y.~Zhu, Z.~Chen, and J.~Luo.
\newblock Anatomy-aware 3d human pose estimation with bone-based pose
  decomposition.
\newblock {\em IEEE Transactions on Circuits and Systems for Video Technology},
  PP:1--1, 02 2021.

\bibitem{camdisthumanpose3d}
H.~Cho, Y.~Cho, J.~Yu, and J.~Kim.
\newblock Camera distortion-aware 3d human pose estimation in video with
  optimization-based meta-learning.
\newblock In {\em Proceedings of the IEEE/CVF International Conference on
  Computer Vision (ICCV)}, pages 11169--11178, October 2021.

\bibitem{ref7_icip_GRU}
K.~Cho, B.~van Merrienboer, {\c{C}}.~G{\"{u}}l{\c{c}}ehre, F.~Bougares,
  H.~Schwenk, and Y.~Bengio.
\newblock Learning phrase representations using {RNN} encoder-decoder for
  statistical machine translation.
\newblock {\em CoRR}, abs/1406.1078, 2014.

\bibitem{ref16_icip_TCMR}
H.~Choi, G.~Moon, J.~Y. Chang, and K.~M. Lee.
\newblock Beyond static features for temporally consistent 3d human pose and
  shape from a video.
\newblock In {\em Proceedings of the IEEE/CVF Conference on Computer Vision and
  Pattern Recognition (CVPR)}, 2021.

\bibitem{ref8_icip_ViT}
A.~Dosovitskiy, L.~Beyer, A.~Kolesnikov, D.~Weissenborn, X.~Zhai,
  T.~Unterthiner, M.~Dehghani, M.~Minderer, G.~Heigold, S.~Gelly, J.~Uszkoreit,
  and N.~Houlsby.
\newblock An image is worth 16x16 words: Transformers for image recognition at
  scale.
\newblock {\em ICLR}, 2021.

\bibitem{ref12_icip_resnet}
K.~He, X.~Zhang, S.~Ren, and J.~Sun.
\newblock Deep residual learning for image recognition.
\newblock {\em CoRR}, abs/1512.03385, 2015.

\bibitem{ref23_icip_h36m}
C.~Ionescu, D.~Papava, V.~Olaru, and C.~Sminchisescu.
\newblock Human3.6m: Large scale datasets and predictive methods for 3d human
  sensing in natural environments.
\newblock {\em IEEE Transactions on Pattern Analysis and Machine Intelligence},
  36(7):1325--1339, 2014.

\bibitem{ref12_Coherent}
W.~Jiang, N.~Kolotouros, G.~Pavlakos, X.~Zhou, and K.~Daniilidis.
\newblock Coherent reconstruction of multiple humans from a single image.
\newblock In {\em Proceedings of the IEEE/CVF Conference on Computer Vision and
  Pattern Recognition (CVPR)}, June 2020.

\bibitem{ref13_icip_HMR}
A.~Kanazawa, M.~J. Black, D.~W. Jacobs, and J.~Malik.
\newblock End-to-end recovery of human shape and pose.
\newblock In {\em Computer Vision and Pattern Regognition (CVPR)}, 2018.

\bibitem{ref17_icip_HMMR}
A.~Kanazawa, J.~Y. Zhang, P.~Felsen, and J.~Malik.
\newblock Learning 3d human dynamics from video.
\newblock In {\em CVPR}, 2019.

\bibitem{ref24_icip_adam}
D.~P. Kingma and J.~Ba.
\newblock Adam: {A} method for stochastic optimization.
\newblock In {\em 3rd International Conference on Learning Representations
  (ICLR)}, 2015.

\bibitem{ref3_icip_VIBE}
M.~Kocabas, N.~Athanasiou, and M.~J. Black.
\newblock Vibe: Video inference for human body pose and shape estimation.
\newblock In {\em Proceedings of the IEEE/CVF Conference on Computer Vision and
  Pattern Recognition (CVPR)}, 2020.

\bibitem{ref14_icip_SPIN}
N.~Kolotouros, G.~Pavlakos, M.~J. Black, and K.~Daniilidis.
\newblock Learning to reconstruct 3d human pose and shape via model-fitting in
  the loop.
\newblock In {\em Proceedings of the IEEE International Conference on Computer
  Vision}, 2019.

\bibitem{ref18_icip_CMR}
N.~Kolotouros, G.~Pavlakos, and K.~Daniilidis.
\newblock Convolutional mesh regression for single-image human shape
  reconstruction.
\newblock In {\em Proceedings of the IEEE/CVF Conference on Computer Vision and
  Pattern Recognition (CVPR)}, 2019.

\bibitem{ref_appearance}
J.~N. Kundu, M.~Rakesh, V.~Jampani, R.~M. Venkatesh, and R.~V. Babu.
\newblock Appearance consensus driven self-supervised human mesh recovery.
\newblock In {\em Proceedings of the European Conference on Computer Vision
  (ECCV)}, 2020.

\bibitem{ref5_icip_METRO}
K.~Lin, L.~Wang, and Z.~Liu.
\newblock End-to-end human pose and mesh reconstruction with transformers.
\newblock In {\em CVPR}, 2021.

\bibitem{ref12_liu2020attention}
R.~Liu, J.~Shen, H.~Wang, C.~Chen, S.-c. Cheung, and V.~Asari.
\newblock Attention mechanism exploits temporal contexts: Real-time 3d human
  pose reconstruction.
\newblock In {\em Proceedings of the IEEE/CVF Conference on Computer Vision and
  Pattern Recognition}, pages 5064--5073, 2020.

\bibitem{ref1_icip_SMPL}
M.~Loper, N.~Mahmood, J.~Romero, G.~Pons-Moll, and M.~J. Black.
\newblock {SMPL}: A skinned multi-person linear model.
\newblock {\em ACM Trans. Graphics (Proc. SIGGRAPH Asia)}, 34(6):248:1--248:16,
  Oct. 2015.

\bibitem{ref4_icip_MEVA}
Z.~Luo, S.~A. Golestaneh, and K.~M. Kitani.
\newblock 3d human motion estimation via motion compression and refinement.
\newblock In {\em Proceedings of the Asian Conference on Computer Vision
  (ACCV)}, 2020.

\bibitem{ref26_amass}
N.~Mahmood, N.~Ghorbani, N.~F.~Troje, G.~Pons-Moll, and M.~J. Black.
\newblock Amass: Archive of motion capture as surface shapes.
\newblock In {\em The IEEE International Conference on Computer Vision (ICCV)},
  Oct 2019.

\bibitem{ref7_martinez2017simple}
J.~Martinez, R.~Hossain, J.~Romero, and J.~J. Little.
\newblock A simple yet effective baseline for 3d human pose estimation.
\newblock In {\em Proceedings of the IEEE International Conference on Computer
  Vision}, pages 2640--2649, 2017.

\bibitem{ref21_icip_mpii3d}
D.~Mehta, H.~Rhodin, D.~Casas, P.~Fua, O.~Sotnychenko, W.~Xu, and C.~Theobalt.
\newblock Monocular 3d human pose estimation in the wild using improved cnn
  supervision.
\newblock In {\em 3D Vision (3DV), 2017 Fifth International Conference on}.
  IEEE, 2017.

\bibitem{ref15_NeuralBodyFit}
M.~Omran, C.~Lassner, G.~Pons-Moll, P.~V. Gehler, and B.~Schiele.
\newblock Neural body fitting: Unifying deep learning and model-based human
  pose and shape estimation.
\newblock Verona, Italy, 2018.

\bibitem{ref25_icip_star}
A.~A.~A. Osman, T.~Bolkart, and M.~J. Black.
\newblock {STAR}: A sparse trained articulated human body regressor.
\newblock In {\em European Conference on Computer Vision (ECCV)}, 2020.

\bibitem{ref22_icip_smplify-x}
G.~Pavlakos, V.~Choutas, N.~Ghorbani, T.~Bolkart, A.~A.~A. Osman, D.~Tzionas,
  and M.~J. Black.
\newblock Expressive body capture: 3d hands, face, and body from a single
  image.
\newblock In {\em Proceedings IEEE Conf. on Computer Vision and Pattern
  Recognition (CVPR)}, 2019.

\bibitem{ref_texturepose}
G.~Pavlakos, N.~Kolotouros, and K.~Daniilidis.
\newblock Texturepose: Supervising human mesh estimation with texture
  consistency.
\newblock In {\em ICCV}, 2019.

\bibitem{ref_coarse2fine}
G.~Pavlakos, X.~Zhou, K.~G. Derpanis, and K.~Daniilidis.
\newblock Coarse-to-fine volumetric prediction for single-image 3d human pose.
\newblock In {\em Proceedings of the IEEE Conference on Computer Vision and
  Pattern Recognition (CVPR)}, July 2017.

\bibitem{ref9_pavllo20193d}
D.~Pavllo, C.~Feichtenhofer, D.~Grangier, and M.~Auli.
\newblock 3d human pose estimation in video with temporal convolutions and
  semi-supervised training.
\newblock In {\em Proceedings of the IEEE/CVF Conference on Computer Vision and
  Pattern Recognition}, pages 7753--7762, 2019.

\bibitem{ref_delving}
Y.~Rong, Z.~Liu, C.~Li, K.~Cao, and C.~C. Loy.
\newblock Delving deep into hybrid annotations for 3d human recovery in the
  wild.
\newblock In {\em The IEEE International Conference on Computer Vision (ICCV)},
  October 2019.

\bibitem{rw_ref9_Self}
H.-Y. Tung, H.-W. Tung, E.~Yumer, and K.~Fragkiadaki.
\newblock Self-supervised learning of motion capture.
\newblock In I.~Guyon, U.~V. Luxburg, S.~Bengio, H.~Wallach, R.~Fergus,
  S.~Vishwanathan, and R.~Garnett, editors, {\em Advances in Neural Information
  Processing Systems 30}, pages 5236--5246. Curran Associates, Inc., 2017.

\bibitem{rw_ref8_BodyNet}
G.~Varol, D.~Ceylan, B.~Russell, J.~Yang, E.~Yumer, I.~Laptev, and C.~Schmid.
\newblock Bodynet: Volumetric inference of 3d human body shapes.
\newblock In {\em Proceedings of the European Conference on Computer Vision
  (ECCV)}, pages 20--36, 2018.

\bibitem{ref6_icip_Transformer}
A.~Vaswani, N.~Shazeer, N.~Parmar, J.~Uszkoreit, L.~Jones, A.~N. Gomez, L.~u.
  Kaiser, and I.~Polosukhin.
\newblock Attention is all you need.
\newblock In {\em Advances in Neural Information Processing Systems},
  volume~30. Curran Associates, Inc., 2017.

\bibitem{ref9_icip_3dpw}
T.~von Marcard, R.~Henschel, M.~J. Black, B.~Rosenhahn, and G.~Pons-Moll.
\newblock Recovering accurate 3d human pose in the wild using imus and a moving
  camera.
\newblock In {\em Proceedings of the European Conference on Computer Vision
  (ECCV)}, 2018.

\bibitem{ref2_icip_GHUM}
H.~Xu, E.~G. Bazavan, A.~Zanfir, W.~T. Freeman, R.~Sukthankar, and
  C.~Sminchisescu.
\newblock Ghum \& ghuml: Generative 3d human shape and articulated pose models.
\newblock In {\em Proceedings of the IEEE/CVF Conference on Computer Vision and
  Pattern Recognition}, 2020.

\bibitem{ref17_DeepKinematics_CVPR2020}
J.~Xu, Z.~Yu, B.~Ni, J.~Yang, X.~Yang, and W.~Zhang.
\newblock Deep kinematics analysis for monocular 3d human pose estimation.
\newblock In {\em Proceedings of the IEEE/CVF Conference on Computer Vision and
  Pattern Recognition (CVPR)}, June 2020.

\bibitem{ref20_icip_penn}
W.~Zhang, M.~Zhu, and K.~G. Derpanis.
\newblock From actemes to action: A strongly-supervised representation for
  detailed action understanding.
\newblock In {\em Proceedings of the IEEE International Conference on Computer
  Vision (ICCV)}, 2013.

\bibitem{ref8_zhao2019semantic}
L.~Zhao, X.~Peng, Y.~Tian, M.~Kapadia, and D.~N. Metaxas.
\newblock Semantic graph convolutional networks for 3d human pose regression.
\newblock In {\em Proceedings of the IEEE/CVF Conference on Computer Vision and
  Pattern Recognition}, pages 3425--3435, 2019.

\end{thebibliography}
